\documentclass[10pt, a4paper]{article}
\usepackage{lrec}
\usepackage{multibib}
\newcites{languageresource}{Language Resources}
\usepackage{graphicx}
\usepackage{tabularx}
\usepackage{soul}
\usepackage{latexsym}
\usepackage{amsmath}
\usepackage{url}
\usepackage{bbm}
\usepackage{todonotes}
\usepackage{booktabs}

\usepackage{epstopdf}
\usepackage[latin1]{inputenc}

\usepackage{hyperref}
\usepackage{xstring}

\DeclareMathOperator{\tf}{tf}
\DeclareMathOperator{\tfidf}{tf-idf}
\DeclareMathOperator{\coverage}{coverage}
\DeclareMathOperator{\dist}{dist}
\DeclareMathOperator{\SPD}{SPD}
\DeclareMathOperator{\pscore}{\mathit{p}-score}
\DeclareMathOperator{\gold}{gold}
\DeclareMathOperator{\LCH}{LCH}
\DeclareMathOperator{\hscore}{\mathit{h}-score}
\DeclareMathOperator{\hpcscore}{\mathit{hpc}-score}
\DeclareMathOperator{\hpcavg}{\mathit{hpc}-avg}

\title{\textbf{Improving Hypernymy Extraction with Distributional Semantic Classes}}

\name{Alexander Panchenko$^{1*}$, Dmitry Ustalov$^{2,3*}$, Stefano Faralli$^3$, Simone P. Ponzetto$^3$, Chris Biemann$^1$}

\address{$^*$ these authors contributed equally \\
         $^1$ University of Hamburg, Department of Informatics, Language Technology Group, Germany \\
         $^2$ University of Mannheim, School of Business Informatics and Mathematics, Data and Web Science Group, Germany\\
         $^3$  Ural Federal University, Institute of Natural Sciences and Mathematics, Russia\\ 
\url{{panchenko,biemann}@informatik.uni-hamburg.de} \\
\url{{dmitry,stefano,simone}@informatik.uni-mannheim.de}\\ }

\abstract{
In this paper, we show how distributionally-induced semantic classes can be helpful  for extracting hypernyms. We  present methods for inducing sense-aware semantic classes using distributional semantics and using these induced semantic classes for filtering noisy hypernymy relations. Denoising of hypernyms is performed by labeling each semantic class with its hypernyms. On the one hand, this allows us to filter out wrong extractions using the global structure of  distributionally similar senses. On the other hand, we infer missing hypernyms via label propagation to cluster terms. We conduct a large-scale crowdsourcing study showing that processing of automatically extracted hypernyms using our approach improves the quality of the hypernymy extraction in terms of both precision and recall. Furthermore, we show the utility of our method in the domain taxonomy induction task, achieving the state-of-the-art results on a SemEval'16 task on taxonomy induction.  \\ \newline  }

\begin{document}

\setlength{\abovedisplayskip}{3pt}
\setlength{\belowdisplayskip}{3pt}

\maketitleabstract

\section{Introduction}

Hypernyms are useful in various applications, such as question answering~\cite{Zhou:13}, query expansion~\cite{gong2005web},  and semantic role labelling~\cite{shi2005putting}
as they can help to overcome sparsity of statistical models. Hypernyms are also the building blocks for learning taxonomies from text~\cite{bordea2016semeval}. Consider the following sentence: ``This caf\'e serves fresh \textit{mangosteen} juice''. Here the infrequent word ``mangosteen'' may be poorly represented or even absent in the vocabulary of a statistical model, yet it can be substituted by lexical items with better representations, which carry close meaning, such as its hypernym ``fruit'' or one of its close co-hyponyms, e.g. ``mango''. 

\begin{figure}[ht]
  \centering
  \includegraphics[width=.45\textwidth]{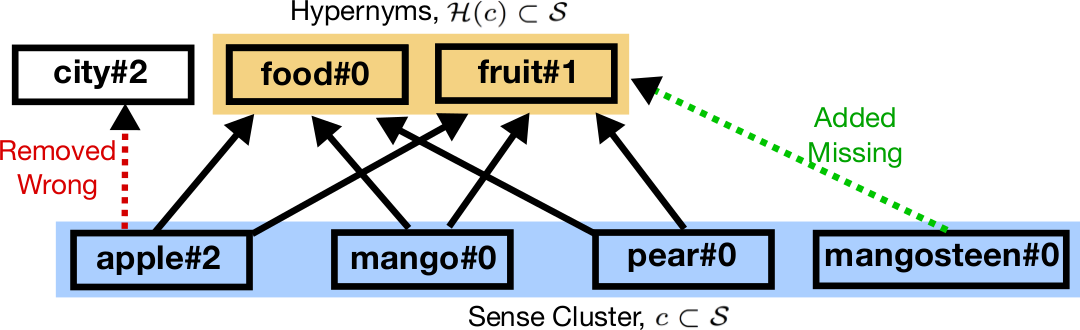}
  \caption{Our approach performs post-processing of hypernymy relations using distributionally induced semantic classes, represented by clusters of induced word senses labeled with noisy hypernyms. The word postfix, such as \texttt{\#1}, is an ID of an induced sense. The wrong hypernyms outside the cluster labels are removed, while the missing ones not present in the noisy database of hypernyms are added. }
  \label{fig:cosetbinary}
\end{figure}

Currently available approaches to hypernymy extraction focus on the acquisition of individual binary hypernymy relations~\cite{hearst1992automatic,snow2004learning,weeds2014learning,shwartz-goldberg-dagan:2016:P16-1,glavavs-ponzetto:2017:EMNLP2017}. Frequencies of the extracted relations usually follow a power-law, with a long tail of noisy extractions containing rare words. We propose a method that performs post-processing of such noisy binary hypernyms using distributional semantics, cf. Figure~\ref{fig:cosetbinary}. Namely, we use the observation that distributionally related words are often are co-hyponyms~\cite{wandmacher2005semantic,Heylen:08} and operationalize it to perform filtering of noisy relations by finding dense graphs composed of both hypernyms and co-hyponyms.  
 
The contribution of the paper is an unsupervised method for post-processing of noisy hypernymy relations based on  clustering of graphs of word senses induced from text. The idea to use distributional semantics to find hypernyms seems natural and has been widely used. However, the existing methods used distributional, yet \textit{sense-unaware} and \textit{local} features. We are the first to use \textit{global sense-aware distributional structure} via the induced semantic classes to improve hypernymy extraction. The implementation of our method and the induced language resources (distributional semantic classes and cleansed hypernymy relations) are available online.\footnote{\url{https://github.com/uhh-lt/mangosteen}}

\begin{figure*}[ht]
  \centering
  \includegraphics[width=.89\textwidth]{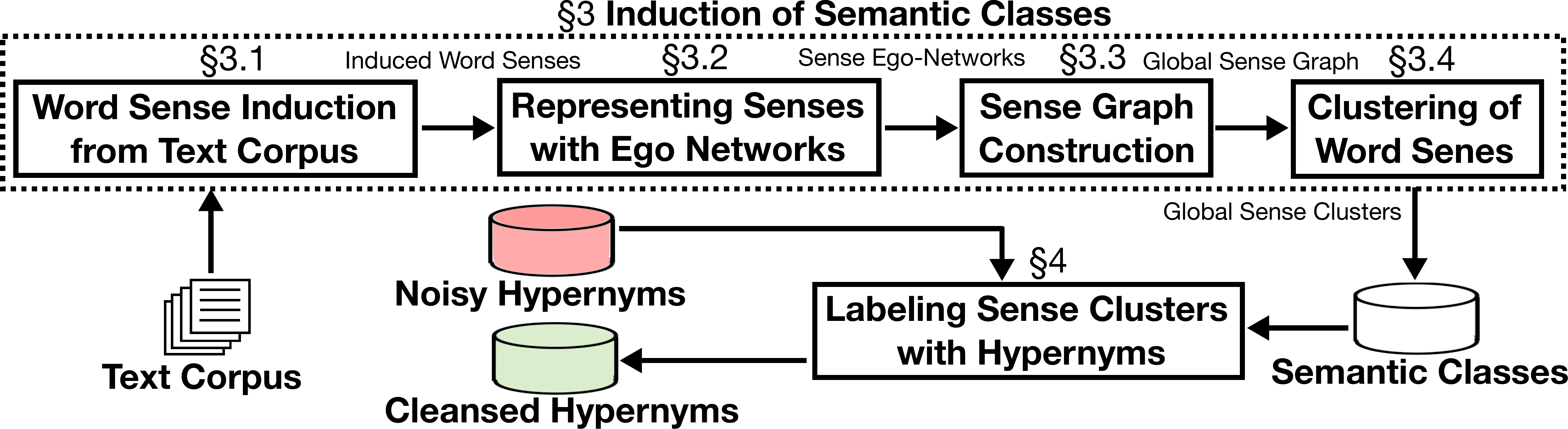}
  \caption{Outline of our approach: sense-aware distributional semantic classes are induced from a text corpus  and then used to filter noisy hypernyms database (e.g. extracted by an external method from a text corpus). }
  \label{fig:outline}
\end{figure*}

\section{Related Work}
\label{sec:related}

\subsection{Extraction of Hypernyms}

In her pioneering work, \newcite{hearst1992automatic} proposed to extract hypernyms based on lexical-syntactic patterns from text. \newcite{snow2004learning} learned such patterns automatically based on a set of hyponym-hypernym pairs. \newcite{pantel2006espresso} presented another approach for weakly supervised extraction of similar extraction patterns. These approaches use some training pairs of hypernyms to bootstrap the pattern discovery process. For instance, \newcite{sang2007extracting} used web snippets as a corpus for extraction of hypernyms. More recent approaches exploring the use of distributional word representations for extraction of hypernyms and co-hyponyms include~\cite{roller2014inclusive,weeds2014learning,necsulescu2015reading,vylomova-EtAl:2016:P16-1}. They rely on two distributional vectors to characterize a relation between two words, e.g. on the basis of the difference of such vectors or their concatenation. \newcite{Levy:15} discovered a tendency to lexical memorization of such approaches, hampering their generalization to other domains.

\newcite{fu2014learning} relied on an alternative approach where a projection matrix is learned, which transforms a distributional vector of a hyponym to the vector of its hypernym. \newcite{ustalov-EtAl:2017:EACLshort} improved this method by adding regularizers in the model that take into account negative training samples and the asymmetric nature of the hypernyms. 

Recent approaches to hypernym extraction focused on learning \textit{supervised} models based on a combination of syntactic patterns and distributional features~\cite{shwartz-goldberg-dagan:2016:P16-1}. Note that while methods, such as \cite{mirkin-dagan-geffet:2006:POS} and \cite{shwartz-goldberg-dagan:2016:P16-1} use distributional features for extraction of hypernyms, in contrast to our method, they do not take into account word senses and    global distributional structure. 

\newcite{seitner2016large} performed extraction of hypernyms from the web-scale Common Crawl\footnote{\url{http://www.commoncrawl.org}} text corpus to ensure high lexical coverage. In our experiments, we use this web-scale database of noisy hypernyms, as the large-scale repository of automatically extracted hypernyms to date. 

\subsection{Taxonomy and Ontology Learning}

 Most relevant in the context of automatic construction of lexical resource are methods for building resources from  text~\cite{caraballo1999automatic,Biemann:05,cimiano2006ontology,bordea2015semeval,velardi2013ontolearn} as opposed to methods that automatically construct resources from semi-structured data~\cite{auer2007dbpedia,navigli2012babelnet} or using crowdsourcing~\cite{biemann2013creating,Braslavski:16:gwc}. 

Our representation differs from the global hierarchy of words as constructed e.g. by \cite{berant2011global,Faralli:16}, as we are grouping many lexical items into a labeled sense cluster as opposed to organizing them in deep hierarchies. \newcite{kozareva2013tailoring} proposed a taxonomy induction method based on extraction of hypernyms using the doubly-anchored lexical patterns. Graph algorithms are used to induce a proper tree from the binary relations harvested from text. 

\subsection{Induction of Semantic Classes}
This line of research starts with \cite{Lin2001}, where sets of similar words are clustered into concepts. While this approach performs a hard clustering and does not label clusters, these drawbacks are addressed in \cite{Pantel2002}, where words can belong to several clusters, thus representing senses, and in \cite{Pantel2004}, where authors aggregate hypernyms per cluster, which come from Hearst patterns. The main difference to our approach is that we explicitly represent senses both in clusters and in their hypernym labels, which enables us to connect our sense clusters into a global taxonomic structure. Consequently, we are the first to use semantic classes to improve hypernymy extraction. 

\newcite{ustalov-panchenko-biemann:2017:Long} proposed a synset induction approach based on global clustering of word senses. The authors used the graph constructed of dictionary synonyms, while we use distributionally-induced graphs of senses.

\begin{table*}
\centering
\footnotesize
\begin{tabular}{l|p{9cm}|p{3.5cm}} 
\bf ID: Word Sense, $s \in \mathcal{S}$ & \bf  Local Sense Cluster: Related Senses, $\mathcal{N}(s) \subset \mathcal{S}$ & \bf   Hypernyms, $\mathcal{H}(s) \subset \mathcal{S}$ \\
\toprule
 mango\#0 &  peach\#1, grape\#0, plum\#0, apple\#0, apricot\#0, watermelon\#1, banana\#1, coconut\#0, pear\#0, fig\#0, melon\#0,  \textbf{mangosteen\#0}, ... & fruit\#0, food\#0, ... \\
 
\midrule
apple\#0 & mango\#0, pineapple\#0, banana\#1, melon\#0, grape\#0, peach\#1, watermelon\#1, apricot\#0, cranberry\#0, pumpkin\#0, \textbf{mangosteen\#0}, ... & fruit\#0, crop\#0,  ... \\

\midrule
Java\#1 & C\#4, Python\#3, Apache\#3, Ruby\#6, Flash\#1, C++\#0, SQL\#0, ASP\#2, Visual Basic\#1, CSS\#0, Delphi\#2, MySQL\#0, Excel\#0, Pascal\#0, ... & programming language\#3, language\#0, ... \\

\midrule
Python\#3 & PHP\#0, Pascal\#0, Java\#1, SQL\#0, Visual Basic\#1, C++\#0, JavaScript\#0, Apache\#3, Haskell\#5, .NET\#1, C\#4, SQL Server\#0, ... & language\#0, technology\#0, ... \\

\end{tabular}
\caption{
Sample induced sense inventory entries  representing ``fruits'' and ``programming language'' senses. Each word sense $s$ is represented with a list of related senses $\mathcal{N}(s)$ and the list of hypernyms $\mathcal{H}(s)$. The hypernyms can be used as human-interpretable sense labels of the sense clusters. One sense $s$, such as ``apple\#0'', can appear in multiple entries.
}
\label{tab:pcz}
\end{table*}

\begin{table*}[ht]
\centering
\footnotesize
\begin{tabular}{l|p{4.5in}|p{1.5in}} 
\bf ID &  \bf Global Sense Cluster: Semantic Class, $c \subset \mathcal{S}$ & \bf Hypernyms, $\mathcal{H}(c) \subset \mathcal{S}$  \\ 

\toprule

1 & peach\#1, banana\#1, pineapple\#0, berry\#0, blackberry\#0, grapefruit\#0, strawberry\#0, blueberry\#0, fruit\#0, grape\#0, melon\#0, orange\#0, pear\#0, plum\#0, raspberry\#0, watermelon\#0, apple\#0, apricot\#0, watermelon\#0, pumpkin\#0, berry\#0, \textbf{mangosteen\#0}, ...  & vegetable\#0, fruit\#0, crop\#0, ingredient\#0, food\#0, $\cdot$ \\ 

\midrule

2  & C\#4, Basic\#2, Haskell\#5, Flash\#1, Java\#1, Pascal\#0, Ruby\#6, PHP\#0, Ada\#1, Oracle\#3, Python\#3, Apache\#3, Visual Basic\#1, ASP\#2, Delphi\#2, SQL Server\#0, CSS\#0, AJAX\#0, JavaScript\#0, SQL Server\#0, Apache\#3, Delphi\#2, Haskell\#5, .NET\#1, CSS\#0, ... & programming language\#3, technology\#0, language\#0, format\#2, app\#0
\end{tabular}
\caption{Sample of the induced  sense clusters representing ``fruits'' and ``programming language'' semantic classes. Similarly to the induced word senses, the semantic classes are labeled with hypernyms. In contrast to the induced word senses, which represent a local clustering of word senses (related to a given word) semantic classes represent a global sense clustering of word senses. One sense $c$, such as ``apple\#0'', can appear only in a single cluster.
}
\label{tab:coset}
\end{table*}

\section{Unsupervised Induction of Distributional Sense-Aware Semantic Classes}
\label{sec:clustering}

As illustrated in Figure~\ref{fig:outline}, our method   induces a sense inventory from a text corpus using the method of \cite{Faralli:16,biemann2018framework}, and clusters these senses.

Sample word senses from the induced sense inventory are presented in Table~\ref{tab:pcz}. The difference of the induced sense inventory from the sense clustering presented in Table~\ref{tab:coset} is that word senses in the induced resource are specific to a given target word, e.g. words ``apple'' and ``mango'' have distinct ``fruit'' senses, represented by a list of related senses. On the other hand, sense clusters represent a global and not a local clustering of senses, i.e. the ``apple'' in the ``fruit'' sense can be a member of only one cluster. This is similar to WordNet, where one sense can only belong to a single synset. Below we describe each step of our method.

\subsection{Word Sense Induction from a Text Corpus   }
\label{sec:induction}

 Each word sense $s$ in the induced sense inventory $\mathcal{S}$ is represented by a list of neighbors $\mathcal{N}(s)$, see Table \ref{tab:pcz} for an example.
 Extraction of this network is performed using the method of~\newcite{Faralli:16} and involves three steps: (1) building a distributional thesaurus, i.e. a graph of related ambiguous terms~\cite{Biemann:13}; (2) word sense induction via clustering of ego networks~\cite{widdows2002graph,everett2005ego} of related words using the Chinese Whispers graph clustering algorithm~\cite{Biemann:06}; (3) disambiguation of related words and hypernyms. The word sense inventory used in our experiment\footnote{The input and output datasets are available for download at \url{https://doi.org/10.5281/zenodo.1174041}}
was extracted from a 9.3 billion tokens corpus, which is a concatenation of  Wikipedia\footnote{\url{https://doi.org/10.5281/zenodo.229904}}, ukWac~\cite{ferraresi2008introducing}, LCC~\cite{LCC} and Gigaword~\cite{graff2003english}. Note that analogous graphs of senses can be obtained using word sense embeddings, see \cite{neelakantanefficient,bartunov2015breaking}. Similarly to any other distributional word graph, the induced sense inventory sense network is scale-free, cf.~\cite{steyvers2005large}. Our experiments show that a global clustering of this network can lead to a discovery of giant components, which are useless in our context as they represent no semantic class. To overcome this problem, we re-build the sense network as described below.     

\subsection{Representing Senses with Ego Networks}

To  perform a global clustering of senses, we represent each induced sense $s$ by a second-order \textit{ego network}~\cite{everett2005ego}. An ego network is a graph  consisting of all related senses $\mathcal{R}(s)$ of the ego sense $s$ reachable via a path of length one or two, defined as: 
\begin{equation}
\{s_j : (s_j \in \mathcal{N}(s)) \vee (s_i \in \mathcal{N}(s) \wedge s_j \in \mathcal{N}(s_i))\}.
\end{equation}

Each edge weight $\mathcal{W}_s(s_i, s_j)$ between two senses is taken from the induced sense inventory network~\cite{Faralli:16} and is equal to a distributional semantic relatedness score between $s_i$ and $s_j$.

\begin{figure}[ht]
  \centering
  \includegraphics[width=.5\textwidth]{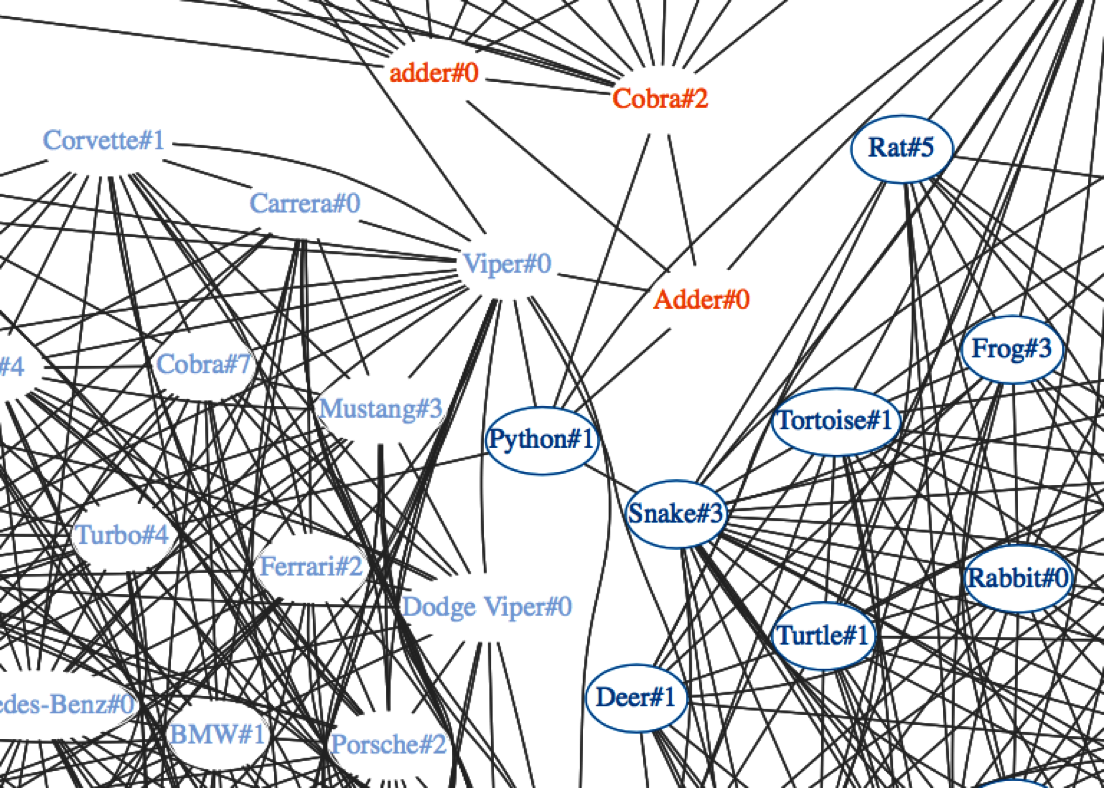}
  \caption{An example of a non-coherent ego network  of the automatically induced sense \texttt{Python\#1}, representing the ``animal" sense. We prune it to remove terms not relevant to the animal sense. }
  \label{fig:ego-network}
\end{figure}

Senses in the induced sense inventory may contain a mixture of different senses introducing noise in a global clustering: cf. Figure~\ref{fig:ego-network}, where  ``Python'' in the animal sense is related to both car and snake senses.
To minimize the impact of the word sense induction errors, we filter out ego networks with a highly segmented structure. Namely, we cluster each ego network with the Chinese Whispers algorithm and discard networks for which the cluster containing the target sense $s$ contains less than 80\% nodes of the respective network to ensure semantic coherence inside the word groups. Besides, all nodes of a network not appearing in the cluster containing the ego sense $s$ are also discarded. 

\subsection{Global Sense Graph Construction}

The goal of this step is to merge ego networks of individual senses constructed at the previous step into a global graph. We compute weights of the edges of the global graph by counting the number of co-occurrences of the same edge in different networks:
\begin{equation}
  \mathcal{W}(s_i, s_j) = \sum_{s \in \mathcal{S}} \mathcal{W}_s(s_i, s_j)\text{.}
\end{equation}

For filtering out noisy edges, we remove all edges with the weight less than a threshold $t$. Finally, we apply the function $E(w)$ that re-scales edge weights. We tested identity function (count) and the natural logarithm (log):
\begin{equation}
\mathcal{W}(s_i, s_j) =
  \begin{cases}
  E(\mathcal{W}(s_i, s_j)) & \text{if } \mathcal{W}(s_i, s_j) \geq T\text{,} \\
  0                        & \text{otherwise.}
  \end{cases}
\end{equation}

\subsection{Clustering of Word Senses}

\begin{figure}[ht]
  \centering
  \includegraphics[width=.5\textwidth]{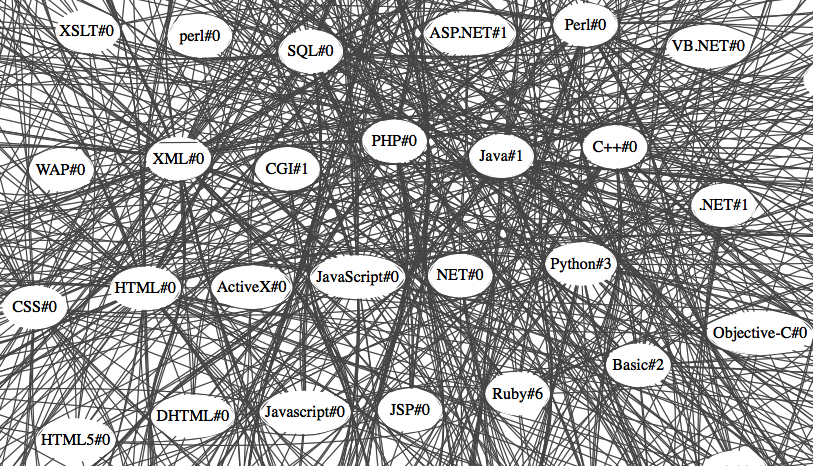}
  \caption{Senses referring to programming languages co-occur in global sense cluster entries, resulting in a densely connected set of co-hyponyms. }
  \label{fig:cluster-programming}
\end{figure}

The core of our method is the induction of semantic classes by clustering the global graph of word senses. We use the Chinese Whispers algorithm to make every sense appear only in one cluster $c$. Results of the algorithm are groups of strongly related word senses that represent different concepts (cf. \figurename~\ref{fig:cluster-programming}). 
Hypernymy is by definition a relation between nouns. Thus optionally, we remove all single-word senses that do not correspond to nouns using the Pattern library~\cite{DeSmedt:12}. This optional mode is configured by the boolean parameter $N$. 

We use two clustering versions in our experiments: the \textit{fine-grained} model clusters 208,871 induced word senses into 1,870 semantic classes, and the \textit{coarse-grained} model that groups 18,028 word senses into 734 semantic classes. To find optimal parameters of our method, we compare the induced labeled sense clusters to lexical semantic knowledge from WordNet~3.1~\cite{Fellbaum:98}
and BabelNet~3.7~\cite{navigli2012babelnet}.

\section{Denoising Hypernyms using the Induced Distributional Semantic Classes}
\label{sec:labeling}

By labeling the induced semantic classes with hypernyms we can thereby remove wrong ones or add those that are missing as illustrated in Figure \ref{fig:cosetbinary}. Each sense cluster is labeled with the noisy input hypernyms, where the labels are the common hypernyms of the cluster word (cf. Table~\ref{tab:coset}). Hypernyms that label no sense cluster are filtered out. In addition, new hypernyms can be generated as a result of labeling. Additional hypernyms are discovered by propagating cluster labels to the rare words without hypernyms, e.g. ``mangosteen'' in Figure~\ref{fig:cosetbinary}. For labeling we used the $\tfidf$ weighting. Hypernyms that appear in many senses $s$ are weighted down: 
\begin{equation}
\tfidf(h) = \sum_{s \in c} \mathcal{H}(s) \cdot \log \frac{\left\vert\mathcal{S}\right\vert}{\left\vert{}h \in \mathcal{H}(s) : \forall s \in \mathcal{S}\right\vert}\text{,}
\end{equation}
where $\sum_{s \in c} \mathcal{H}(s)$ is a sum of weights for all hypernyms for each sense $s$, per each cluster $c$.

We label each sense cluster  $c$ with its top five hypernyms $\mathcal{H}(c)$. Each hypernym is disambiguated using the method of \newcite{Faralli:16}. Namely, we calculate the cosine similarity between the context (the current sense cluster) and the induced senses (local clusters of the ambiguous word).

Distributional representations of rare words, such as ``mangosteen" can be less precise than those of frequent words. However, co-occurrence of a hyponym and a hypernym in a single sentence is not required in our approach, while it is the case for the path-based  hypernymy extraction methods.

\begin{table*}[ht]
\footnotesize
\centering
\begin{tabular}{p{1.25cm}|p{1.96cm}|p{1.4cm}|p{1.4cm}|p{1.6cm}|p{1.4cm}|p{1.4cm}|p{1.4cm}|p{1.4cm}}
 & \textbf{Min. sense co-occurrences}, $t$ & \textbf{Edge weight}, $E$ & \textbf{Only nouns}, $N$ & \textbf{Hypernym weight}, $H$ & \textbf{Number of clusters} & \textbf{Number of senses} & $\hpcavg$, \textbf{WordNet} & $\hpcavg$, \textbf{BabelNet} \\ \toprule

coarse-gr. & \underline{$100$} & \underline{log}   & \underline{yes} & \underline{tf-idf} & \underline{$734$}  &  \underline{$18\,028$} & 
\underline{$\mathbf{0.092}$} & \underline{$\mathbf{0.304}$} \\

 & $100$ & log   & no  & tf-idf & $763$  &  $27\,149$ & $0.090$ & $0.303$ \\
 & $100$ & count & no  & tf-idf & $765$  &  $27\,149$ & $0.089$ & $0.302$ \\
 & $100$ & log   & no  & tf     & $784$  &  $27\,149$ & $0.090$ & $0.300$ \\
 & $100$ & count & yes & tf     & $733$  &  $18\,028$ & $\mathbf{0.092}$ & $0.299$ \\
 & $100$ & count & no  & tf     & $772$  &  $27\,149$ & $0.089$ & $0.297$ \\
 & $100$ & count & yes & tf-idf & $732$  &  $18\,028$ & $0.091$ & $0.295$ \\
 & $100$ & log   & yes & tf     & $726$  &  $18\,028$ & $0.088$ & $0.293$ \\ \midrule

fine-gr. & \underline{$0$}   & \underline{count} & \underline{no}  & \underline{tf-idf} & \underline{$1870$} & \underline{$208\,871$} & \underline{$\mathbf{0.041}$} &  \underline{$\mathbf{0.279}$} \\

 & $0$   & count & no  & tf     & $1877$ & $208\,871$ & $\mathbf{0.041}$ & $0.278$ \\
 & $0$   & count & yes & tf     & $2070$ & $144\,336$ & $0.037$ & $0.240$ \\
 & $0$   & count & yes & tf-idf & $2080$ & $144\,336$ & $0.038$ & $0.240$ \\
 & $0$   & log   & yes & tf-idf & $4709$ & $144\,336$ & $0.027$ & $0.138$ \\
 & $0$   & log   & yes & tf     & $4679$ & $144\,336$ & $0.027$ & $0.136$ \\
 & $0$   & log   & no  & tf-idf & $5960$ & $208\,871$ & $0.035$ & $0.127$ \\
 & $0$   & log   & no  & tf     & $5905$ & $208\,871$ & $0.036$ & $0.126$ 
\end{tabular}
\caption{Performance of different configurations of the hypernymy labeled global sense clusters in terms of their similarity to WordNet/BabelNet. The results are sorted by performance on BabelNet dataset, the best values in each section are boldfaced. The two underlined configurations are respectively the best \textit{coarse-grained} and \textit{fine-grained} grained semantic class models used in all experiments. The coarse grained model contains less semantic classes, but they tend to be more consistent than those of the fine-grained model, which contains more senses and classes. }
\label{tab:results}
\end{table*}

\section{Finding an Optimal Configuration of Meta Parameters of the Method}
\label{sec:params}

The approach consists of several sequential stages, as depicted in Figure~\ref{fig:outline}, with each stage having a few meta parameters. This study is designed to find promising combinations of these meta parameters.  In this section, we propose several metrics which aim at finding an optimal configuration of all these meta parameters jointly. In particular, to compare different configurations of our approach, we compare the labeled sense clusters to WordNet~3.1~\cite{Fellbaum:98} and BabelNet~3.7~\cite{navigli2012babelnet}. The assumption is that the optimal model contains lexical semantic knowledge similar to the knowledge in the lexical resources.  To implement the evaluation metrics we used the NLTK library~\cite{Bird:09} and the BabelNet Java~API.\footnote{\url{http://www.babelnet.org}}

\subsection{Metrics Quantifying Goodness of Fit of the Induced Structures to  Lexical Resources}
To summarize various aspects of the lexical resource, we propose a score that is maximized if  labeled sense clusters are generated directly from a lexical resource:
\begin{equation}
  \hpcscore(c) = \frac{\hscore(c) + 1}{\pscore(c) + 1} \cdot \coverage(c)\text{.}
\end{equation}

$\pscore(c)$ quantifies the plausibility of the sense cluster $c$. It  reflects the distance of co-hyponyms in a lexical resource:
\begin{equation}
  \pscore(c) = \frac{1}{\left\vert{}c\right\vert} \sum^{\left\vert{}c\right\vert}_{i=1} \sum^i_{j=1} \dist(w_i, w_j)\text{.}
\end{equation}

The lower the $\pscore$ is, the closer the hyponyms are located in the gold standard resource. For each pair of distinct lemmas $(w_i, w_j)$ in the cluster of co-hyponyms $c$, we search for the minimal shortest path distance (SPD) between the synsets corresponding to each word in the pair, i.e. $S(w_i)$ is the set of synsets having the $w_i$ lemma and $S(w_j)$ is the similar set with respect to the $w_j$ lemma:
\begin{equation}
  \dist(w_i, w_j) = \min_{\substack{s' \in S(w_i),\\s'' \in S(w_j)}} \SPD(s', s'')\text{.}
\end{equation}

$\hscore(c)$ quantifies plausibility of the hypernyms $\mathcal{H}(c)$ of a sense cluster $c$ measuring the precision of extracted hypernyms:
\begin{equation}
  \hscore(c) = \frac{\left\vert\mathcal{H}(c) \cap \gold(c)\right\vert}{\left\vert\mathcal{H}(c)\right\vert}\text{.}
\end{equation}

The $\gold(c)$ is composed of the lowest common hypernyms (LCH) in the lexical resource for each pair of lemmas in the sense cluster $c$:
\begin{equation}
  \gold(c) = \bigcup_{\substack{w_i \in c,\\w_j \in c}} \bigcup_{\substack{s' \in S(w_i),\\s'' \in S(w_j)}} \{\LCH(s', s'')\}\text{.}
\end{equation}

$\coverage(c)$ quantifies how well cluster words are represented in the gold standard resource. Thus, errors in poorly represented clusters are discounted via coverage. Coverage is the fraction of the lemmas appearing both in the cluster $c$ and in the vocabulary of the resource $\mathcal{V}$:
\begin{equation}
 \coverage(c) = \frac{\left\vert{}c \cap \mathcal{V}\right\vert}{\left\vert{}c\right\vert}.
\end{equation}
The total score used to rank various configurations of our approach averages $\hpcscore$ scores for all induced sense clusters:
\begin{equation}
  \hpcavg = \frac{1}{\left\vert{}\mathcal{C}\right\vert} \sum_{c \in \mathcal{C}} \hpcscore(c)\text{.}
\end{equation}

\begin{figure}
  \centering
  \includegraphics[width=.4\textwidth]{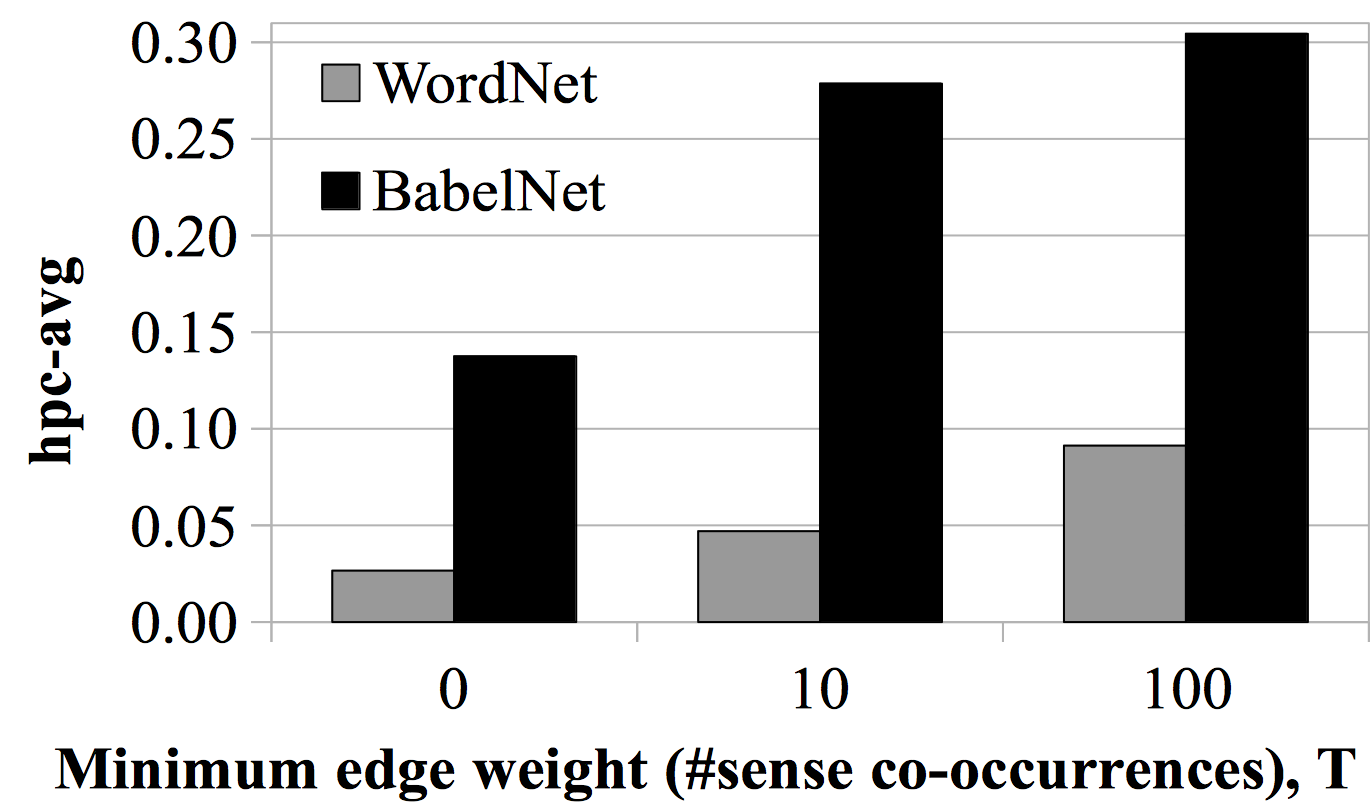}
  \caption{Impact of the min. edge weight $t$. }
  \label{fig:threshold}
\end{figure}

\subsection{Results}

Meta parameter search results based on the comparison to WordNet and BabelNet are provided in Figure~\ref{fig:threshold} and \tablename~\ref{tab:results}. The minimal edge weight $t$ trades off between the size of the resulting resource (number of words and senses) and its similarity to the gold lexical resources. The higher the threshold, the fewer nodes remain in the graph, yet these remaining nodes form densely interlinked communities. For $t$ of 100, each pair of senses in the graph is observed in at least 100 ego networks. Secondly, for the unpruned model ($t=$ 0), edge weights based on counts worked better than logarithmic weights. However, when pruned ($t >$ 0), logarithmic edge weighting shows better results. Thirdly, the $\tfidf$ weights proved to yield consistent improvements over the basic $\tf$ weighting. For the pruned model, the variation in scores across different configurations is small as the underlying graphs are of high quality, while for the unpruned model the choice of parameters has much more impact as the sense graphs are noisier.  

We selected the best-performed configuration according to BabelNet ($\hpcavg$ of 0.304), which also is the second best configuration according to WordNet ($\hpcavg$ of 0.092). This model is based on the edge threshold $t$ of 100, logarithmic weights of edges contains only nouns and hypernyms ranked according to $\tfidf$. Note also that, the best-unpruned model ($t=0$) has  BabelNet $\hpcavg$  of 0.279, which is only 10\% lower than the best model, yet the unpruned model has an order of magnitude larger vocabulary and a more fine-grained representation (734 vs. 1,870 clusters). Thus, if coverage is important, the unpruned model is recommended. In the remainder of this paper, we continue with the first model listed in Table~\ref{tab:results} and evaluate it in the following experiments.

\section{Evaluation}

To evaluate our approach, we conduct two intrinsic evaluations and one extrinsic evaluation. The first experiment aims to estimate the fraction of spurious sense clusters, the second one evaluates the quality of the post-processed hypernyms. Finally, we evaluate the induced semantic classes in application to the taxonomy induction task.

\subsection{Experiment 1: Plausibility of the Induced Semantic Classes}

Comparison to gold standard resources allows us to gauge the relative performances of various configurations of our method. To measure the absolute quality of the best configuration selected in the previous section, we rely on microtask-based crowdsourcing with CrowdFlower\footnote{\url{https://www.crowdflower.com}}. 

\subsubsection{Task Design}

\begin{figure}[t]
  \centering
  \includegraphics[width=.5\textwidth]{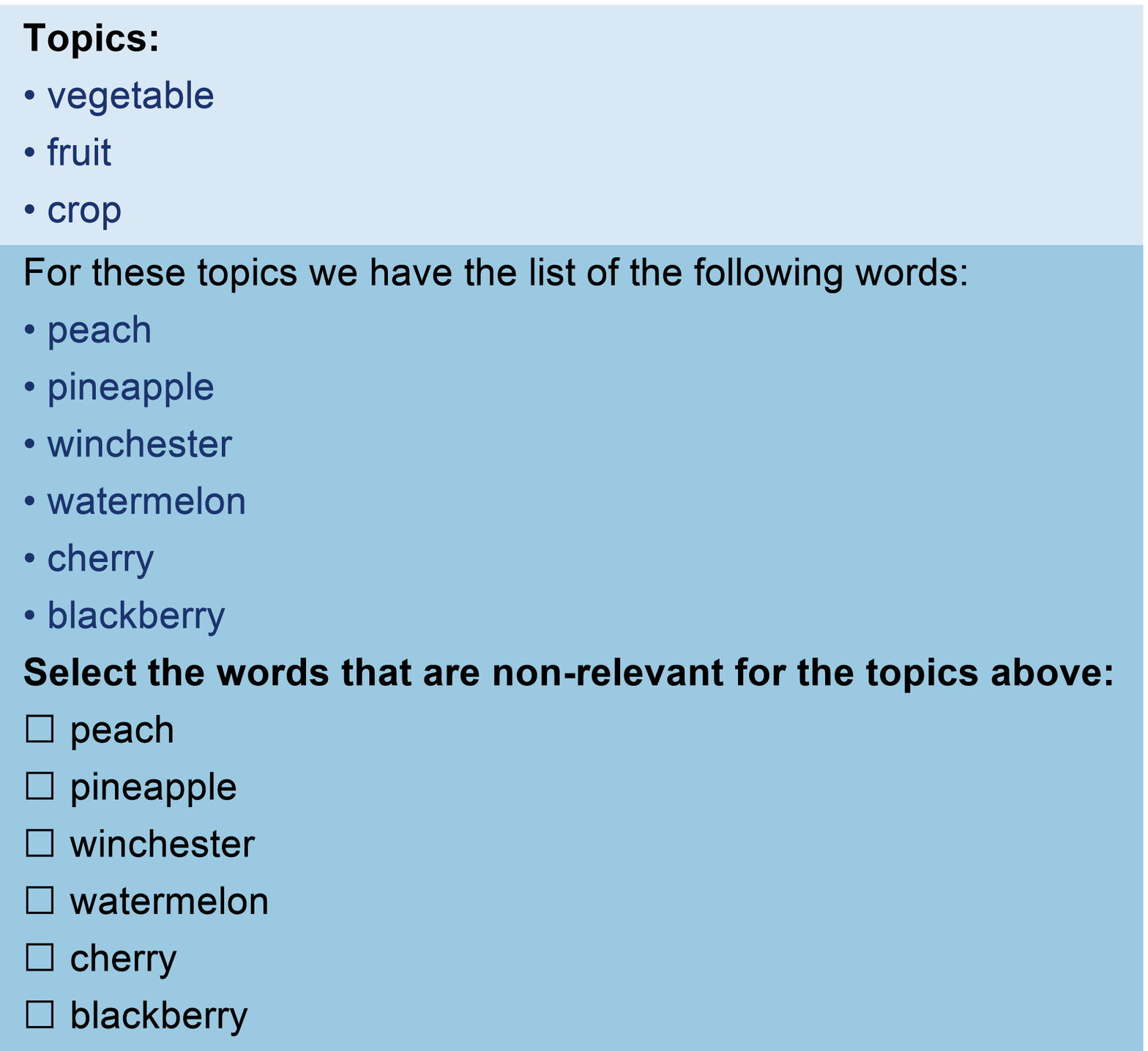}
  \caption{Layout of the sense cluster evaluation crowdsourcing task, the entry ``winchester'' is the intruder.}
  \label{fig:cluster-hit}
\end{figure}

We used two crowdsourcing tasks based on word intruder detection~\cite{Chang:09} to measure how humans perceive the extracted lexical-semantic structures. Namely, the tasks are designed to evaluate the quality of the extracted sense clusters and their labels. The input form presented to an annotator is illustrated in \figurename~\ref{fig:cluster-hit}.
A crowdworker is asked to identify words that do not match the context represented by words from a sense cluster or its label. To generate an intruder, following the original design of~\newcite{Chang:09}, we select a random word from a cluster and replace it with a word of similar frequency that does not belong to any cluster (bias here is low as the evaluated model contains 27,149 out of  313,841 induced word senses). In both tasks, the workers have been provided with concise instructions and test questions. 

\subsubsection{Evaluation Metrics}

We compute two metrics on the basis on annotation results: (1) \textit{accuracy} is the fraction of tasks where annotators correctly identified the intruder, thus the words from the cluster are consistent; (2) \textit{badness} is the fraction of tasks for which non-intruder words were selected. In this experiment, we assume that it is easy to identify the intruder in a correct sense cluster and difficult in a noisy, implausible sense cluster. We compute \textit{accuracy} as the fraction of tasks where annotators correctly identified the intruder, thus the words from the cluster are consistent.

\begin{table}
\footnotesize
\centering

\begin{tabular}{l|c|c|c}
\textbf{} & \textbf{Accuracy} & \textbf{Badness} & \textbf{Randolph $\kappa$} \\ \toprule
Sense clusters, $c$ & $0.859$ & $0.248$ & $0.739$ \\
Hyper. labels, $\mathcal{H}(c)$  & $0.919$ & $0.208$ & $0.705$ \\
\end{tabular}

\caption{Plausibility of the sense clusters according to human judgments via an intruder detection experiment for the coarse-grained semantic class model.}
\label{tab:exp2-results}
\end{table}

\begin{table*}
\footnotesize
\centering

\begin{tabular}{l|c|c|c}
 & \textbf{Precision} & \textbf{Recall} & \textbf{F-score} \\ \toprule
Original hypernymy relations extracted from the Common Crawl corpus~\cite{seitner2016large} & $0.475$ & $0.546$ & $0.508$ \\
Enhanced hypernyms with the \textit{coarse-grained} semantic classes   & $\mathbf{0.541}$ & $\mathbf{0.679}$ & $\mathbf{0.602}$ \\
\end{tabular}

\caption{Results of post-processing of a noisy hypernymy database with our approach, evaluated using human judgements.}
\label{tab:exp3-results}
\end{table*}

\subsubsection{Results}


Table~\ref{tab:exp2-results} summarizes the results of the intruder detection experiment. Overall, 68 annotators provided 2,035 judgments about the quality of sense clusters. Regarding hypernyms, 98 annotators provided 2,245 judgments. The majority of the induced semantic classes and their labels are highly plausible according to human judgments: the accuracy of the sense clusters based on the intruder detection is 0.859 (agreement of 87\%), while the accuracy of hypernyms is 0.919 (agreement of 85\%). The Randolph $\kappa$ of respectively 0.739 and 0.705 indicates substantial inter-observer agreement~\cite{randolph2005free}. 

According to the feedback mechanism of the CrowdFlower, the co-hyponymy task received a 4.0 out of 5.0 rating, while the hypernymy task received a 4.4 out of 5.0 rating. The crowdworkers show a \textit{substantial} agreement according to Randolph $\kappa$ coefficient computed 0.739 for the cluster evaluation task and 0.705 for the hypernym evaluation task.

Major sources of errors for crowdworkers are rare words and entities. While clusters with well-known entities, such as ``Richard Nixon'' and ``Windows Vista'' are correctly labeled, examples of other less-known named entities, e.g. cricket players, are sometimes wrongly labeled as implausible. Another source of errors during crowdsourcing were wrongly assigned hypernyms: in rare cases, sense clusters are labeled with hypernyms like ``thing'' or ``object'' that are too generic even under  $\tfidf$ weighting.

\subsection{Experiment 2: Improving Binary Hypernymy Relations}

In this experiment, we test whether our post-processing based on the semantic class improves the quality of hypernymy relations (cf. Figure~\ref{fig:outline}).

\subsubsection{Generation of Binary Hypernyms.}
We evaluated the best coarse-grained model identified in the first experiment 
($t$ of 100). Each sense cluster of this model is split into the set $H_{cluster}$ of binary hypernyms, as illustrated in Figure~\ref{fig:cosetbinary}. Overall, we gathered 85,290 hypernym relations for 17,058 unique hyponyms. Next, we gathered the set $H_{orig}$ of 75,486 original hypernyms for exactly the same 17,058 hyponyms. For each word from the sense cluster we looked up top five hypernyms under the best ones when sorting them by extraction frequency from the hypernym relation database of~\newcite{seitner2016large} as in our model each sense cluster is labeled with five hypernyms from the same database. The database of \newcite{seitner2016large} is extracted  using lexical patterns. Note that any other method for extraction of binary hypernyms can be used at this point, e.g.~\cite{weeds2014learning,roller2014inclusive,shwartz-goldberg-dagan:2016:P16-1,glavavs-ponzetto:2017:EMNLP2017}. For the comparison, we gathered up to five hypernyms for each word, using (1) the most frequent hypernym relations from \cite{seitner2016large} vs. (2)  the cluster labeling method as described above. 

\subsubsection{Task Design}

\begin{figure}[t]
  \centering
  \includegraphics[width=.375\textwidth]{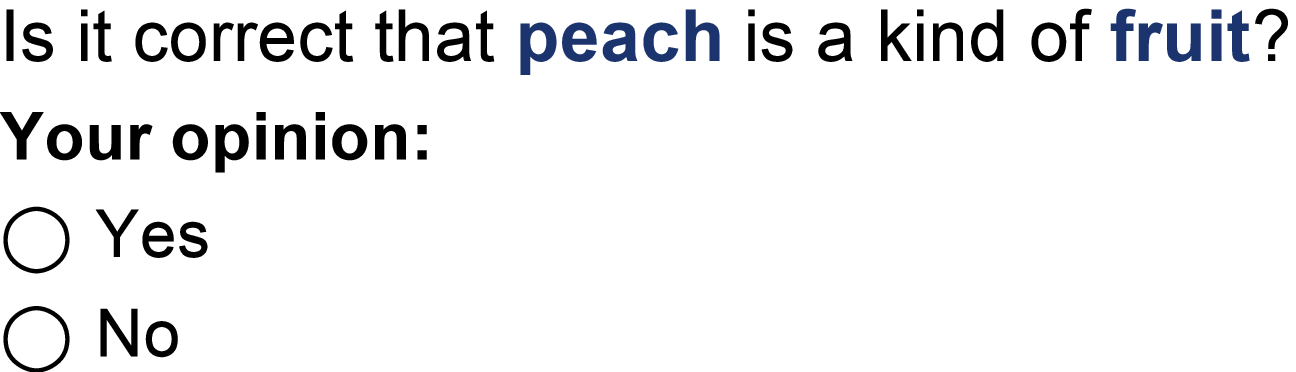}
  \caption{Layout of the hypernymy annotation task.}
  \label{fig:crowdhyper}
\end{figure}

We drew a random sample of 4,870 relations using lexical split by hyponyms. All relations from $H_{cluster}$ and $H_{orig}$ of one hyponym were included in the sample. These relations were subsequently annotated by human judges using crowdsourcing. We asked crowdworkers to provide a binary judgment about the correctness of each hypernymy relation as illustrated in Figure~\ref{fig:crowdhyper}.

\subsubsection{Results} 

Overall, 298 annotators completed 4,870 unique tasks each labeled 6.9 times on average, resulting in a total of 33,719 binary human judgments about hypernyms. We obtained a \textit{fair} agreement among annotators of 0.548 in terms of the Randolph $\kappa$~\cite{Meyer:14}. Since CrowdFlower reports a confidence for each answer, we selected $N=3$ most confident answers per pair and aggregated them using weighted majority voting. The ties were broken pessimistically, i.e. by treating a hypernym as irrelevant. Results for $N\in{3,5,6}$ varied less than by 0.002 in terms of F-score. The task received the rating of a 4.4 out of 5.0 according to the annotator's feedback mechanism. 

Table~\ref{tab:exp3-results} presents results of the experiment. Since each pair received a binary score, we calculated Precision, Recall, and F-measure of two compared methods. Our denoising method improves the quality of the original hypernyms by a large margin both in terms of precision and recall, leading to an overall improvement of 10  F-score points. The improvements of recall are due to the fact that to label a cluster of co-hyponyms it is sufficient to lookup hypernyms for only a fraction of words in the clusters. However, binary relations will be generated between all cluster hypernyms and the cluster words potentially generating hypernyms missing in the input database. For instance, a cluster of fruits can contain common entries like ``apple'' and ``mango'' which ensure labeling it with the word ``fruit''. Rare words in the same cluster, like ``mangosteen'', which have no hypernyms in the original resource due to the sparsity of the pattern-based approach, will also obtain the hypernym ``fruit'' as they are  distributionally related to frequent words with reliable hypernym relations, cf. Figure~\ref{fig:cosetbinary}. We also observed this effect frequently with clusters of named entities, like cricket players. Improvements in precision are due to filtering of wrong extractions, which are different for different words and thus top hypernyms of a cluster contain only hypernyms confirmed by several co-hyponyms. 

Finally, note that all previous hypernymy extraction methods output binary relations between undisambiguated words (cf. Section~\ref{sec:related}). Therefore, our approach could be used to improve results of other state-of-the-art hypernymy extraction approaches, such as HypeNET~\cite{shwartz-goldberg-dagan:2016:P16-1}.

\begin{table*}
\footnotesize
\centering
\begin{tabular}{l|p{1.3cm}|p{1.3cm}|p{1.5cm}|p{1.5cm}|p{1.3cm}|p{1.6cm}}
\textbf{System / Domain, Dataset} & \textbf{Food, WordNet} & \textbf{Science, WordNet}& \textbf{Food, Combined} & \textbf{Science, Combined} & \textbf{Science, Eurovoc} & \textbf{Environment, Eurovoc} \\ \toprule

WordNet & $1.0000$ & $1.0000$ & $0.5870$ & $0.5760$ & $0.6243$ & n.a. \\ \midrule

Baseline & $0.0022$ & $0.0016$ & $0.0019$ & $0.0163$ & $0.0056$ & $0.0000$ \\
JUNLP & $0.1925$ & $0.0494$ & $0.2608$ & $0.1774$ & $0.1373$ & $0.0814$ \\
NUIG-UNLP & n.a. & $0.0027$ & n.a. & $0.0090$ & $0.1517$ & $0.0007$ \\
QASSIT & n.a. & $0.2255$ & n.a. & $0.5757$ & $0.3893$ & $0.4349$ \\
TAXI & $0.3260$ & $0.2255$ & $0.2021$ & $0.3634$ & $0.3893$ & $0.2384$ \\
USAAR & $0.0021$ & $0.0008$ & $0.0000$ & $0.0020$ & $0.0023$ & $0.0007$ \\ \midrule

Semantic Classes (fine-grained) & $0.4540$ & $0.4181$ & $0.5147$ & $0.6359$ &  $\mathbf{0.5831}$ & $0.5600$ \\
Semantic Classes (coarse-grained) & $\mathbf{0.4774}$ & $\mathbf{0.5927}$ & $\mathbf{0.5799}$ & $\mathbf{0.6539}$ & $0.5515$ & $\mathbf{0.6326}$ 
\end{tabular}
\caption{Comparison of the our taxonomy induction method on the SemEval 2016 Task 13 on Taxonomy Extraction Evaluation for English in terms of cumulative Fowlkes\&Mallows measure (F\&M).}
\label{tab:taxo}
\end{table*}

\begin{table}[ht]
\footnotesize
\centering
\begin{tabular}{l|p{0.9cm}|p{1.2cm}|p{1.3cm}|p{1.3cm}}
\textbf{Domain} & \textbf{\#Seeds words} & \textbf{\#Expand. words} & \textbf{\#Clusters}, fine-gr. & \textbf{\#Clusters}, coarse-gr.  \\ \toprule
Food & $2\,834$ & $3\,047$ & $29$ & $21$ \\
Science & $806$ & $1\,137$ & $73$ & $35$ \\
Environ. & $261$ & $909$ & $111$ & $39$ \\
\end{tabular}
\caption{Summary of the domain-specific sense clusters.}
\label{tab:domain-cosets}
\end{table}

\subsection{Experiment 3: Improving Domain Taxonomy Induction}

In this section, we show how the labeled semantic classes can be used for induction of domain taxonomies.

\subsubsection{SemEval 2016 Task 13}

We use the taxonomy extraction evaluation dataset by \newcite{bordea2016semeval}, featuring gold standard taxonomies for three domains (Food, Science, Environment) and four languages (English, Dutch, French, and Italian) on the basis of existing lexical resources, such as WordNet and Eurovoc~\cite{steinberger2006jrc}.\footnote{\url{http://eurovoc.europa.eu}} Participants were supposed to build a taxonomy provided a vocabulary of a domain. Since our other experiments were conducted on English, we used the English part of the task. The evaluation is based on the Fowlkes\&Mallows Measure (F\&M), a cumulative measure of the similarity of both taxonomies~\cite{velardi2013ontolearn}. 

\subsubsection{Taxonomy Induction using Semantic Classes}

Our method for taxonomy induction takes as input a vocabulary of the domain and outputs a taxonomy of the domain. The method consists of three steps: (1) retrieving sense clusters relevant to the target domain; (2) generation of binary relations though a Cartesian product of words in a sense cluster and its labels; (3) attaching disconnected components to the root (the name of the domain). We retrieve domain-specific senses for each domain of the SemEval datasets by a  lexical filtering. First, we build an extended lexicon of each domain on the basis of the seed vocabulary of the domain provided in the SemEval dataset. Namely, for each seed term, we retrieve all semantically similar terms. To filter out noisy expansions, related terms are added to the expanded vocabulary only if there are at least $k=5$ common terms between the seed vocabulary and the list of related terms. Second, we retrieve all sense clusters that contain at least one term from the expanded vocabulary among its sense clusters or hypernyms. Table~\ref{tab:domain-cosets} summarizes results of this domain filtering. 
After, we generate binary hypernymy relations by linking every word in the semantic class to each hypernymy label as shown in Figure~\ref{fig:cosetbinary}. Finally, we link roots of each disconnected components to the root of the taxonomy, e.g. ``food'' for the Food domain. Note that this step was used by SemEval participants, e.g. in the TAXI system~\cite{panchenko-EtAl:2016:SemEval}.

\subsubsection{Results}

Table~\ref{tab:taxo} presents results of the taxonomy extraction experiment. We evaluated two best models of our method: a \textit{coarse} and a \textit{fine} grained clusterings featuring respectively 734 and 1870 semantic classes identified in Section~\ref{sec:params} with different levels of pruning: $t \in \{0, 100\}$. As one can observe, our model based on the labeled sense clusters  significantly outperforms the substring-based baseline and all participating system by a large margin on all domains. For the ``Science (Eurovoc)'' and ``Food'' domains our method yields results comparable to WordNet while remaining unsupervised and knowledge-free. Besides, for the ``Science'' domain our method outperforms WordNet, indicating on the high quality of the extracted lexical semantic knowledge. Overall, the \textit{coarse-grained} more pruned model yielded better results as compared to \textit{fine-grained} un-pruned model for all domains but ``Science (Eurovoc)''.

\section{Conclusion}

In this paper, we presented an unsupervised method for the induction of sense-aware semantic classes using distributional semantics and graph clustering and showed how these can be used for post-processing of noisy hypernymy databases extracted from text.
We determined optimal parameters of our approach by a comparison to existing lexical-semantic networks.
To evaluate our approach, we performed three experiments. A large-scale crowdsourcing study indicated a high plausibility of extracted semantic classes according to human judgment. Besides, we demonstrated that our approach helps to improve precision and recall of a hypernymy extraction method. Finally, we showed how the proposed semantic classes can be used to improve domain taxonomy induction from text. 

While we have demonstrated the utility of our approach for hypernym extraction and taxonomy induction, we believe that the induced semantic classes can be useful in other tasks. For instance, in \cite{panchenko-EtAl:2017:EMNLP2017Demos} these semantic classes were used as an inventory for word sense disambiguation to deal with out of vocabulary words.

\section{Acknowledgements}

This research was supported by Deutscher Akademischer Austauschdienst (DAAD), Deutsche For\-schungs\-gemeinschaft (DFG) under the project "Joining Ontologies and Semantics Induced from Text" (JOIN-T), by the RFBR under the project no.~16-37-00354~mol\_a, and by the Ministry of Education and Science of the Russian Federation Agreement no.~02.A03.21.0006. We are grateful to three anonymous reviewers for their helpful comments. Finally, we are grateful to Dirk Johann\ss en for providing feedback on an early version of this paper.

\section{Bibliographical References}
\label{main:ref}

\bibliographystyle{lrec}
\bibliography{lrec}


\end{document}